\documentclass[sigconf,nonacm]{acmart}

\usepackage{booktabs}
\usepackage{amsmath}
\usepackage{graphicx}
\usepackage{xcolor}
\usepackage{verbatim}
\usepackage{placeins}
\usepackage{tabularx}
\usepackage{makecell}
\usepackage{array}
\usepackage{algorithm}
\usepackage{algpseudocode}
\usepackage{fontawesome5}
\usepackage[most]{tcolorbox}
\usepackage{listings}
\usepackage{pdflscape}

\newcolumntype{Y}{>{\centering\arraybackslash}X}

\setcopyright{none}

\definecolor{promptblue}{HTML}{31579B}
\definecolor{promptbg}{HTML}{F6F7F9}
\definecolor{promptborder}{HTML}{B8BEC6}

\newtcblisting{rawprompt}[1]{
    enhanced,
    breakable,
    listing only,
    width=0.88\textwidth,
    center,
    colback=promptbg,
    colframe=promptborder,
    colbacktitle=promptblue,
    coltitle=white,
    title={#1},
    fonttitle=\bfseries\ttfamily,
    boxrule=0.8pt,
    arc=2mm,
    left=4mm,
    right=4mm,
    top=2mm,
    bottom=2mm,
    listing options={
        basicstyle=\ttfamily\small,
        breaklines=true,
        columns=fullflexible,
        keepspaces=true,
        showstringspaces=false
    }
}

\AtBeginDocument{
  \providecommand\BibTeX{{
    \normalfont
    B\kern-0.5em{\scshape i\kern-0.25em b}
    \kern-0.8em\TeX
  }}
}

\begin{document}

\title{Your AI, On a Dial: Controlling Investment Bias in LLMs with a Single Neuron}

\author{Sahong Park}
\email{justin4396@hufs.ac.kr}
\affiliation{%
  \institution{Hankuk University of Foreign Studies}
  \city{Yongin}
  \country{Republic of Korea}
}

\author{Suhwan Park}
\email{suhwan@unist.ac.kr}
\affiliation{%
  \institution{UNIST}
  \city{Ulsan}
  \country{Republic of Korea}
}

\author{Hoyoung Lee}
\email{hoyounglee@unist.ac.kr}
\affiliation{%
  \institution{UNIST}
  \city{Ulsan}
  \country{Republic of Korea}
}

\author{Gakyung Kwon}
\email{eileenk@hufs.ac.kr}
\affiliation{%
  \institution{Hankuk University of Foreign Studies}
  \city{Yongin}
  \country{Republic of Korea}
}

\author{Wonbin Ahn}
\email{wonbin.ahn@lgresearch.ai}
\affiliation{%
  \institution{LG AI Research}
  \city{Seoul}
  \country{Republic of Korea}
}

\author{Jaewon Choi}
\email{jaewonch@hanwha.com}
\affiliation{%
  \institution{Hanwha Life}
  \city{Seoul}
  \country{Republic of Korea}
}

\author{Alejandro Lopez-Lira}
\email{alejandro.lopez-lira@warrington.ufl.edu}
\affiliation{%
  \institution{University of Florida}
  \city{Gainesville}
  \state{FL}
  \country{United States}
}

\author{Yoon Kim}
\email{yoonkim@mit.edu}
\affiliation{%
  \institution{Massachusetts Institute of Technology}
  \city{Cambridge}
  \state{MA}
  \country{United States}
}

\author{Chanyeol Choi}
\email{jacobchoi@linqalpha.com}
\affiliation{%
  \institution{LinqAlpha}
  \city{New York}
  \state{NY}
  \country{United States}
}

\author{Hyeongwoo Kong}
\authornote{Corresponding authors: Hyeongwoo Kong and Yongjae Lee.\\
Preprint. Under review.}
\email{gho3283@hufs.ac.kr}
\affiliation{%
  \institution{Hankuk University of Foreign Studies}
  \city{Yongin}
  \country{Republic of Korea}
}

\author{Yongjae Lee}
\authornotemark[1]
\email{yongjaelee@unist.ac.kr}
\affiliation{%
  \institution{UNIST}
  \city{Ulsan}
  \country{Republic of Korea}
}
\affiliation{%
  \institution{LinqAlpha}
  \city{New York}
  \state{NY}
  \country{United States}
}

\renewcommand{\shortauthors}{Park et al.}

\begin{abstract}
Large language models (LLMs) are increasingly used in investment
decision-making, yet prior work shows that they exhibit systematic,
model-specific investment preferences. We study whether a model's overall
investment stance can be calibrated to a specified direction and strength.
We introduce an investment-bias dial, an inference-time intervention on a
single neuron that continuously adjusts a model-level decision prior---its
overall tendency toward buying or selling---without targeting specific firms
or investment attributes. Using matched positive and negative evidence, we
evaluate five open-weight LLMs and find that the dial produces monotonic
changes in investment stance without modifying prompts or model parameters.
At the response level, the dial shifts both investment decisions and the
evidential emphasis of generated rationales under identical inputs. In an
agentic retrieval setting, the dial also changes what information the model
searches for, which evidence it selects, and which evidence is reflected in
its final analysis. In a long-context evaluation, the dial maintains stable
stance control as context length increases, whereas a matched system-prompt
instruction progressively attenuates. We further show that changes in the
dial propagate to security rankings and downstream portfolio composition in
an exploratory backtest. Overall, our results show that an LLM's aggregate
investment stance can be calibrated toward a specified target at inference
time.
\end{abstract}


\maketitle

\section{Introduction}
\label{sec:intro}

Large Language Models (LLMs) are playing an increasingly prominent role in financial decision-making, supporting tasks from market analysis and investment research to portfolio construction \cite{hwang2025dinn,lee2025blacklitterman,xiao2024tradingagents}. As their influence grows, it becomes important to understand not only the decisions they produce, but also the underlying views that shape those decisions. LLMs are not necessarily neutral financial decision-makers: they can exhibit systematic, domain-specific biases that affect how evidence is interpreted and how investment judgments are formed \cite{kong2026evaluating}.

Prior work documents model-specific preferences across sector, firm-size, and momentum dimensions, some of which persist even in the face of conflicting evidence \cite{lee2025bias}. In investment settings, however, there may be no universally appropriate stance; the desired stance can depend on the investor's objective or mandate. We therefore frame the problem not as eliminating bias, but as giving users control over the model's overall investment stance---both its direction and its strength. Existing methods make this adjustment difficult. Prompting can be sensitive to wording and context, while fine-tuning commits the model to a fixed target and must be repeated when that target changes \cite{bini2026behavioral,gao2026debiasing}.

\begin{figure*}[t]
    \centering
    \includegraphics[width=\textwidth]{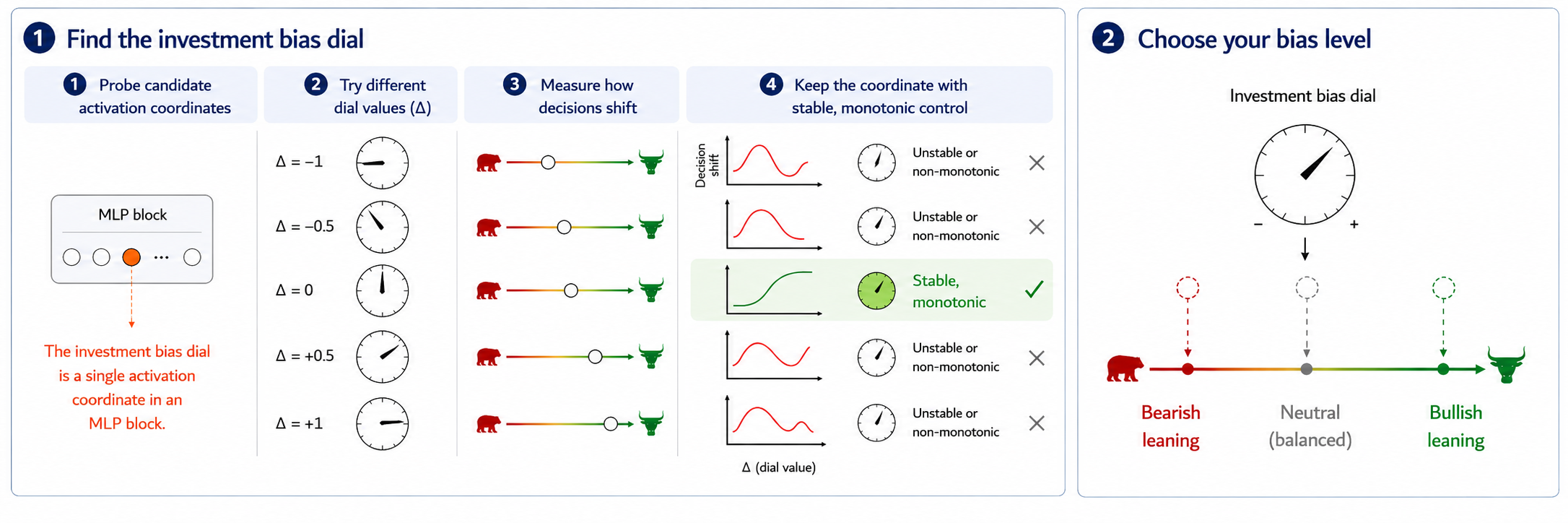}
    \caption{
    Overview of the proposed investment-bias dial.
    We intervene on candidate activation coordinates in an MLP block and vary
    the intervention coefficient $\Delta$ to measure how the model's investment
    decisions change. Candidate coordinates are first screened for decision relevance,
    and feasible candidates are ranked by held-out calibration error.
    The selected coordinate is then calibrated to target bearish,
    neutral, or bullish investment preferences.
    }
    \label{fig1:overview}
\end{figure*}

We introduce an \textbf{investment-bias dial}, an inference-time intervention on a single neuron. Turning the dial changes a model-level decision prior---its general tendency to buy or sell---rather than targeting particular securities or investment attributes. The dial therefore provides continuous control over the model's overall stance without changing the prompt or updating model parameters. Figure~\ref{fig1:overview} summarizes how the neuron is selected and adjusted at inference time.

We address three research questions:

\begin{itemize}
    \item \textbf{RQ1:} Can a single neuron provide monotonic control over an LLM's overall investment stance?
    \item \textbf{RQ2:} Can the dial be calibrated to a target investment stance, including a neutral stance?
    \item \textbf{RQ3:} Does the effect of the investment-bias dial persist and propagate beyond controlled buy--sell decisions to downstream investment workflows?
\end{itemize}

Using a controlled evidence-conflict framework \cite{lee2025bias}, we evaluated the dial on five open-weight LLMs spanning different model families and scales. Across all models, changes in the dial setting produce monotonic shifts in the investment decision prior and enable calibration toward specified targets, including neutrality. Keeping the evidence fixed, the dial also changes both the final investment decision and the information emphasized in the rationale. Across seven general-purpose and finance-specific benchmarks, we
find no evidence of systematic performance degradation. In additional evaluations, the dial affects information
acquisition and evidence use, maintains stance control over long
contexts relative to matched prompting, and propagates to security
rankings and portfolio composition.

Our contributions are threefold. First, we propose the \textbf{investment-bias dial}, a simple yet novel single-neuron intervention for continuously controlling an LLM's overall investment stance at inference time without changing prompts or model parameters. Second, across five open-weight LLMs, we show that the dial produces monotonic shifts in the investment decision prior and can be calibrated to a target stance. Third, we examine the practical implications of the dial for agentic
research behavior, long-context stance control, and downstream
portfolio construction.

\section{Related Work}
\label{sec:related}

\begin{comment}
This section reviews previous work on financial biases in LLMs and methods for modifying model behavior to contextualize our investigation of controllable investment bias.
\end{comment}

\subsection{Financial Biases in Large Language Models}

Recent studies have documented systematic biases in LLM-based financial
reasoning and decision-making. Lee et al.~\cite{lee2025bias} introduce a
controlled evidence-conflict framework for measuring latent investment
preferences and examine whether these preferences persist when models are
presented with contradictory evidence. Across sector, firm-size, and momentum
dimensions, they identify model-specific investment preferences and show that
some are associated with confirmation bias. Related work has examined biases
arising from the presentation and representation of financial information.
Pairwise financial decisions exhibit positional bias, with model choices
varying with the ordering of alternatives and the magnitude of this effect
differing across model scales and prompting conditions
~\cite{dimino2025positional}. A separate study of open-source Qwen models
finds systematic variation in model confidence with firm size, valuation,
risk, and industry sector~\cite{dimino2025representation}. Product-level
analyses likewise document systematic preferences for specific financial
products across multiple asset classes~\cite{zhi2025product}.

Financial biases have also been linked to differences in information
availability and broader behavioral tendencies. Prior work documents
foreign-market bias in financial predictions associated with differences
in information environments~\cite{cao2025foreign}, as well as systematic
behavioral patterns across preference- and belief-based economic and
financial decisions~\cite{bini2026behavioral}. Collectively, these studies
document a range of financial biases associated with latent investment
preferences, option ordering, firm and product characteristics, information
availability, and decision framing. In contrast to prior work that primarily
identifies and characterizes such biases, our work investigates whether a
measured investment preference can be continuously adjusted toward a specified
target through a localized internal intervention.

\subsection{Modifying Model Behavior}

Previous work has sought to modify LLM behavior through prompting,
fine-tuning, and inference-time intervention. In financial decision tasks,
rationality-oriented prompting has been shown to reduce several behavioral
biases~\cite{bini2026behavioral}. Parameter-level approaches have also been
explored: LoRA-based supervised fine-tuning can mitigate extrapolation bias
in financial forecasting and improve out-of-sample predictions
~\cite{gao2026debiasing}. These approaches demonstrate that financial
behavior can be modified, but they rely on either prompt design or parameter
updates rather than exposing a direct internal control for continuously
adjusting the behavior at inference time.

Activation-based methods instead intervene directly on internal
representations during inference. Inference-Time Intervention modifies a
small set of attention-head activations along directions associated with
truthfulness~\cite{li2023iti}. Representation Engineering provides a broader
framework for monitoring and manipulating population-level representations
associated with high-level model behavior~\cite{zou2023repe}, while
Contrastive Activation Addition constructs steering vectors from activation
differences between contrastive examples and adds them during forward
passes~\cite{rimsky2024caa}. Related work has extended activation steering
to social-bias mitigation; FairSteer detects bias-related activation patterns
and conditionally applies debiasing steering vectors during inference
~\cite{li2025fairsteer}. At the same time, steering effects can vary
substantially across inputs and prompting conditions, and indiscriminate
intervention can introduce performance side effects on inputs for which
steering is unnecessary~\cite{tan2024steering,stickland2024steering}.

A complementary line of work studies behavior at the level of individual
neurons. Neuron-level analyses have identified units associated with factual
knowledge and shown that manipulating them can alter factual expression
~\cite{dai2022knowledge}. Similar approaches have localized neurons
associated with social bias and used targeted suppression to mitigate biased
behavior~\cite{liu2024devil}. More recently, single-neuron interventions
have been shown to substantially alter safety-related behavior, including
bypassing safety alignment through the suppression of an individual neuron
~\cite{kazemi2026neuron}. These studies demonstrate that highly localized
internal interventions can affect model outputs, but they do not study
continuous calibration of an investment preference. We therefore examine
whether an additive intervention on a single MLP coordinate can provide
monotonic and calibratable control over an investment-bias score while
limiting changes in other evaluated capabilities.

\section{Preliminaries}
\label{sec:preliminaries}
\subsection{Investment-Bias Measurement}

We adopt the bias elicitation protocol of Lee et al.~\cite{lee2025bias}. Let $\mathcal{S}= \left\{s_1,\ldots,s_{427}\right\}$ denote the ticker universe. For each ticker $s\in\mathcal{S}$, a balanced trial presents equal numbers of matched bullish and bearish evidence items in randomized order and asks the model to choose between \texttt{buy} and \texttt{sell}. Because the evidence is balanced by construction, systematic deviations from an equal number of buy and sell decisions indicate an investment bias.

Let $N^{(s)}_{\mathrm{buy}}$ and $N^{(s)}_{\mathrm{sell}}$ denote the numbers of outputs parsed as buy and sell for ticker $s$, respectively. We define the ticker-level investment-bias score as
\begin{equation}
\pi_s = \frac{N^{(s)}_{\mathrm{buy}} - N^{(s)}_{\mathrm{sell}}}
             {N^{(s)}_{\mathrm{buy}} + N^{(s)}_{\mathrm{sell}}} \in [-1, 1],
\label{eq:ticker-index}
\end{equation}
when the denominator is nonzero. Positive values indicate a buy bias, whereas negative values indicate a sell bias. A value of zero indicates an equal number of parsed buy and sell decisions for the ticker.

Aggregating buy and sell counts across all tickers in $\mathcal{S}$,
we define the model-level investment-bias score as
\begin{equation}
\pi =
\frac{\sum_{s\in\mathcal{S}} N^{(s)}_{\mathrm{buy}}
      -\sum_{s\in\mathcal{S}} N^{(s)}_{\mathrm{sell}}}
     {\sum_{s\in\mathcal{S}} N^{(s)}_{\mathrm{buy}}
      +\sum_{s\in\mathcal{S}} N^{(s)}_{\mathrm{sell}}}.
\label{eq:model-index}
\end{equation}
We use $\pi$ as the primary measure of the model's overall investment bias. Values closer to $1$ indicate a stronger aggregate buy bias, values closer to $-1$ indicate a stronger aggregate sell bias, and $\pi=0$ denotes the neutral point. The exact elicitation prompt and evaluation protocol are provided in
Appendix~\ref{app:prompt}.

\subsection{Neuron Intervention}

Each decoder block contains an MLP whose hidden activations are projected
back to the residual stream through a down-projection. Previous work has
shown that individual MLP neurons can be causally manipulated to alter
model behavior~\cite{dai2022knowledge,kazemi2026neuron}. Following this
neuron-level intervention paradigm, we apply an additive shift to a
single coordinate of the input to the MLP down-projection.

We focus on a single-coordinate intervention to test whether highly
localized internal control is sufficient. Unlike activation-steering
methods that typically apply dense directions such as CAA or RepE, our
intervention modifies only one coordinate of an MLP intermediate
activation.

Let $\ell$ index an MLP block, and let
$\mathbf{h}_{\ell,t} \in \mathbb{R}^{d_{\mathrm{ff}}}$ denote the input
to its down-projection at token position $t$, where $d_{\mathrm{ff}}$ is
the MLP intermediate dimension. For a selected coordinate $n$ and
intervention strength $\Delta \in \mathbb{R}$, we define
\begin{equation}
\widetilde{h}_{\ell,t}[n]
=
h_{\ell,t}[n] + \Delta,
\label{eq:neuron-intervention}
\end{equation}
at every token position, including both prompt and generated tokens
during autoregressive decoding. All other coordinates are left
unchanged. The intervention modifies neither the model parameters nor
the input prompt. We implement the intervention with a pre-hook on the
MLP down-projection.

\section{Method}
\label{sec:method}

\begin{comment}
    This section describes how we identify and calibrate the single-neuron
    investment-bias dial and evaluate whether the intervention preserves general
    and finance-specific model capabilities.
\end{comment}

\paragraph{Models.}
We evaluate five open-weight LLMs spanning different model
families and scales. Table~\ref{tab:models} summarizes their
architectural characteristics.

\begin{table}[h]
\centering
\caption{Open-weight LLMs evaluated in this study.}
\label{tab:models}
\small
\setlength{\tabcolsep}{3pt}

\begin{tabularx}{\columnwidth}{
@{}
l
>{\raggedright\arraybackslash}X
c
c
@{}
}
\toprule
Provider & Model & Params. & Layers \\
\midrule
Alibaba
& Qwen3-8B
& 8.2B
& 36 \\

Meta
& Llama-4-Scout-17B-16E-Instruct
& 17B (109B)
& 48 \\

DeepSeek
& DeepSeek-R1-Distill-Qwen-14B
& 14B
& 48 \\

Google
& Gemma-4-12B-it
& 12B
& 48 \\

Mistral
& Mistral-Small-24B-Instruct-2501
& 24B
& 40 \\
\bottomrule
\end{tabularx}
\end{table}

\subsection{Finding the Investment-Bias Dial}
\label{sec:finding-dial}

% 절차 간소화하여 작성

We select the dial coordinate according to two criteria:
decision relevance and prior controllability.

\paragraph{Decision relevance.}
We first identify MLP coordinates that are locally associated with
the model's buy--sell decision. Let $c=(\ell,n)$ denote coordinate
$n$ in layer $\ell$, and let $x_{\ell,p}^{(i)}[n]$ be its activation
at token position $p$ for ticker $s$. We define the
gradient sensitivity of coordinate $c$ as
\begin{equation}
G_c
=
\left|
\frac{1}{|\mathcal{S}|}
\sum_{s\in\mathcal{S}}
\sum_p
\frac{\partial
\left(
z_{\mathrm{buy}}^{(s)}
-
z_{\mathrm{sell}}^{(s)}
\right)}
{\partial x_{\ell,p}^{(s)}[n]}
\right|.
\label{eq:gradient_sensitivity}
\end{equation}
A larger $G_c$ indicates that a small change in the coordinate is
more strongly associated with the relative preference for buying
over selling. We use this statistic to reduce the full set of MLP
coordinates to a tractable candidate set.

\paragraph{Prior controllability.}
A decision-relevant coordinate is useful as a dial only if varying
its intervention coefficient can reproduce the desired investment
priors. To assess this property, we divide the security universe
into two disjoint subsets, $\mathcal{S}_A$ and $\mathcal{S}_B$.
For each candidate $c$, we measure its transfer curve
$\pi_{c,A}(\Delta)$ on $\mathcal{S}_A$ and estimate the intervention
coefficient
\begin{equation}
\widehat{\Delta}_{c,A}(t)
=
\operatorname{Inv}\!\left(\pi_{c,A},t\right)
\end{equation}
for each target prior $t\in\mathcal{T}$. The same coefficient is
then applied to $\mathcal{S}_B$ without further calibration. We
measure the candidate's ability to reproduce the target prior grid
by
\begin{equation}
\operatorname{RMSE}(c)
=
\sqrt{
\frac{1}{|\mathcal{T}|}
\sum_{t\in\mathcal{T}}
\left[
\pi_{c,B}
\left(
\widehat{\Delta}_{c,A}(t)
\right)
-
t
\right]^2
}.
\label{eq:prior_calibration_error}
\end{equation}
A lower RMSE value indicates that the coordinate provides more
accurate and transferable control over the investment prior across
securities. Algorithm~\ref{alg:dial_finding} summarizes the complete
dial-selection procedure.

\begin{algorithm}[t]
\caption{Selecting the investment-bias dial}
\label{alg:dial_finding}
\begin{algorithmic}[1]
\Require Model $M$, security universe $\mathcal{S}$, target prior grid $\mathcal{T}$
\Ensure Dial coordinate $c^*$ and calibrated settings
\State Compute gradient sensitivity $G_c$ for all MLP coordinates
\State Screen candidates according to $G_c$
\State Split $\mathcal{S}$ into disjoint subsets $\mathcal{S}_A$ and $\mathcal{S}_B$
\For{all candidate coordinates $c$}
    \State Estimate $\widehat{\Delta}_{c,A}(t)$ for each $t\in\mathcal{T}$ using $\mathcal{S}_A$
    \State Compute $\mathrm{RMSE}(c)$ on $\mathcal{S}_B$
\EndFor
\State Select the feasible candidate with the lowest $\mathrm{RMSE}(c)$
\State Re-estimate its calibrated settings on the full universe
\State \Return $c^*$ and its calibrated settings
\end{algorithmic}
\end{algorithm}

\subsection{Evaluating Capability Preservation}
\label{sec:eva_capbility}

We evaluate whether the investment-bias dial preserves capabilities
beyond the targeted investment preference using five general
benchmarks---MMLU \cite{hendrycks2021mmlu}, GSM8K
\cite{cobbe2021gsm8k}, ARC-Challenge (ARC-C) \cite{clark2018arc},
TruthfulQA \cite{lin2022truthfulqa}, and RACE \cite{lai2017race}---and
two finance-specific benchmarks, FinQA \cite{chen2021finqa} and
Financial PhraseBank (FPB) \cite{malo2014good}. All seven benchmarks
are evaluated zero-shot under a fixed custom protocol. Multiple-choice
tasks are scored using next-token logits over option letters, whereas
GSM8K and FinQA use greedy generation followed by numeric parsing. We
use TruthfulQA-MC1, RACE-all, and the 50\%-agreement FPB split. MMLU
uses a fixed 1,000-item sample, while the remaining benchmarks use
their full evaluation splits after validity filtering.

For each model, the unmodified setting, $\Delta=0$, serves as the
baseline. Using the calibration procedure in Section~4.1, we evaluate
settings targeting $\pi\in\{-0.3,0,+0.3\}$, corresponding to
sell-biased, neutral, and buy-biased preferences, respectively; the
calibrated neutral setting $\widehat{\Delta}_{c^*}(0)$ need not
coincide with $\Delta=0$. Performance at each calibrated setting is
compared with the paired $\Delta=0$ baseline. For the six
accuracy-based benchmarks, paired changes are evaluated using the exact
McNemar test \cite{mcnemar1947sampling}, and we report one-sided 95\%
lower confidence bounds on accuracy changes to characterize potential
performance degradation. McNemar $p$-values are adjusted within each
comparison family using the Benjamini--Hochberg procedure
\cite{benjamini1995fdr}. FPB is evaluated using weighted F1 to account
for class imbalance.

\begin{figure*}[t]
    \centering
    \includegraphics[width=\textwidth]{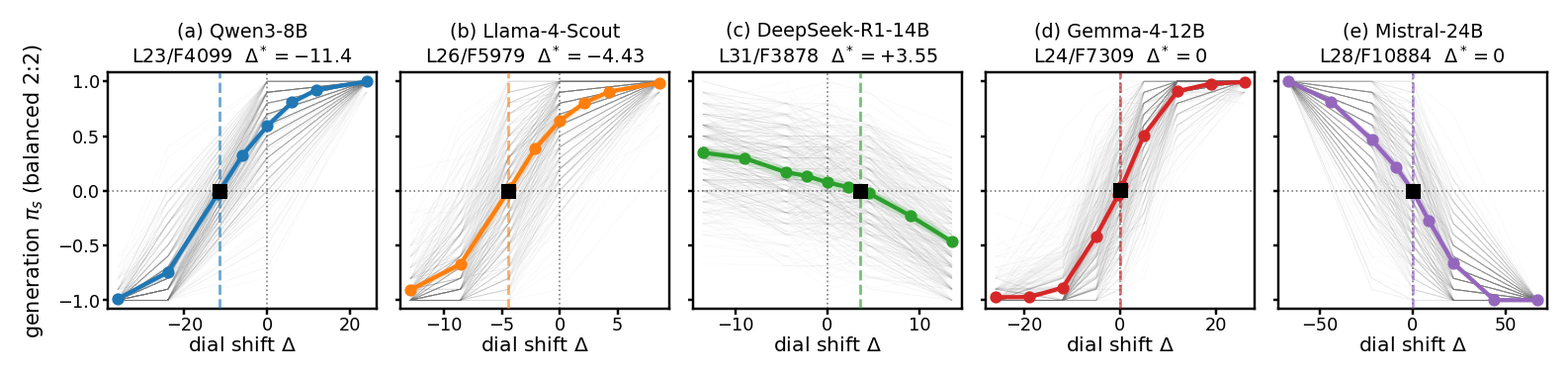}
    \caption{
        Investment-bias responses for the selected coordinates. Colored
        curves show the aggregate investment-bias score $\pi$, and gray
        curves show ticker-level scores $\pi_s$, as functions of intervention
        strength $\Delta$. The black square and vertical dashed line mark the
        calibrated neutral setting $\widehat{\Delta}_{c^*}(0)$. All selected
        coordinates produce monotonic aggregate responses, although their
        directions and reachable ranges differ across models.
    }
    \label{fig:rq1_dial_curves}
\end{figure*}

\section{Results}\label{sec:results}

Additional experiments examine evidence sensitivity under calibration,
prompt-based steering, and dense activation-steering baselines in
Appendices~\ref{app:evidence_sensitivity},
\ref{app:prompt_steering}, and~\ref{app:dense_steering}, respectively.

\subsection{RQ1: Monotonic Investment-Bias Dial}
\label{sec:rq1}

For each model, we select the feasible coordinate with the lowest
RMSE calibration error as the investment-bias dial. As shown in
Figure~\ref{fig:rq1_dial_curves}, the investment-bias score $\pi$ varies
monotonically with the intervention strength $\Delta$ over the evaluated
range for all five models. The ticker-level trajectories exhibit the same
overall ordering, indicating that the aggregate response is not attributable
to a small subset of tickers.

The reachable range is model-dependent. Four models approach the full
interval $[-1,1]$, whereas DeepSeek-R1-14B exhibits a narrower response.
Inspection of its generated responses shows that, even under strong
interventions, the reasoning process can terminate in opposing buy and sell
decisions across prompts. Because $\pi$ aggregates these binary decisions,
such mixed outcomes prevent the score from saturating near either endpoint.
This limits the attainable range but does not affect the monotonic ordering
of the response.

The direction of the dial response is model-specific. In DeepSeek-R1-14B
and Mistral-24B, increasing $\Delta$ shifts $\pi$ in the opposite direction
to that observed for the other models. Because this direction depends on
the model-specific orientation of the selected coordinate, it does not
affect whether the coordinate functions as a dial. We therefore evaluate
dial quality by monotonicity, reachable range, and calibration error rather
than by the sign of the response.

The selected coordinates lie in the middle-to-late portions of their
respective networks. This pattern is consistent with prior layer-wise
analyses. Geva et al.\ \cite{geva2021transformer} found that upper feed-forward layers capture more
semantic patterns than lower layers, while
Dai et al.\ \cite{dai2022knowledge} reported that fact-related knowledge neurons in BERT were
concentrated in later layers. More directly in the
financial domain, Dimino et al.\ \cite{dimino2025positional} localized positional-bias signals in
Qwen2.5 models primarily to mid-to-late transformer layers and identified
recurrent attention heads within these layers. Because these studies examine different
architectures, internal components, and behaviors, the correspondence
should be interpreted as a qualitative similarity in layer-wise
localization rather than evidence of a shared mechanism.

As an illustrative example of the selection procedure, Table~\ref{tab:qwen_candidates}
reports the five feasible Qwen3-8B coordinates with the lowest RMSE.
L23/F4099 attains the lowest RMSE while covering the full target range
and maintaining high valid-decision and parse rates. It is therefore
selected as $c^*$ for Qwen3-8B.

\begin{table}[t]
    \caption{
        Candidate ranking for Qwen3-8B. The five feasible
        coordinates with the lowest RMSE are shown as
        layer/feature indices.
    }
    \vspace{-4pt}
    \label{tab:qwen_candidates}
    \centering
    \small

    \begin{tabular}{lcccc}
        \toprule
        \makecell[c]{Neuron}
        & \makecell[c]{RMSE ($\downarrow$)}
        & \makecell[c]{Range}
        & \makecell[c]{Valid-decision\\rate}
        & \makecell[c]{Parse\\rate} \\
        \midrule
        L23/F4099 & \textbf{0.038} & 2.000 & 0.988 & 1.000 \\
        L25/F9117 & 0.048          & 2.000 & 0.988 & 1.000 \\
        L25/F96   & 0.054          & 2.000 & 0.988 & 1.000 \\
        L26/F9411 & 0.079          & 2.000 & 0.984 & 1.000 \\
        L23/F3183 & 0.108          & 1.727 & 0.988 & 1.000 \\
        \bottomrule
    \end{tabular}

    \vspace{2pt}
    \begin{minipage}{\columnwidth}
        \footnotesize
        \textit{Note.} Valid-decision rate is the proportion of outputs
        yielding a valid \textsc{buy} or \textsc{sell} decision.
    \end{minipage}
\end{table}

\subsection{RQ2: Calibrating the Investment Bias}
\label{sec:RQ2}

Neutrality in investment decision-making is not a universal normative target,
as an appropriate preference may depend on the investor's mandate and market
conditions. We therefore define neutrality operationally as $\pi=0$ under the
balanced-evidence protocol, corresponding to equal aggregate frequencies of
buy and sell decisions, rather than the absence of all investment-related
bias or an optimal investment policy. Table~\ref{tab:neutral_calibration}
reports the baseline investment-bias score, the calibrated intervention
coefficient $\Delta^{*}=\widehat{\Delta}_{c^*}(0)$, and the resulting output
changes. Qwen3-8B and Llama-4-Scout exhibit the largest baseline buy biases,
with calibration changing $29.6\%$ and $32.5\%$ of their decisions,
respectively. DeepSeek-R1-14B begins closer to neutrality, although $21.1\%$
of its decisions change; its positive $\Delta^{*}$ reflects the reversed
dial direction discussed in Section~\ref{sec:rq1}. Gemma-4-12B and
Mistral-24B are already consistent with the neutral target at baseline, so
calibration assigns $\Delta^{*}=0$ and leaves their outputs unchanged.
Mistral-24B's near-neutral baseline is also consistent with the LinqAlpha
Investment Bias Leaderboard~\cite{linqalphaBiasLeaderboard}, where it had
the lowest reported bias score among the listed open-weight models at the
time of access.

\begin{figure}[t]
    \centering
    \includegraphics[width=\columnwidth]
    {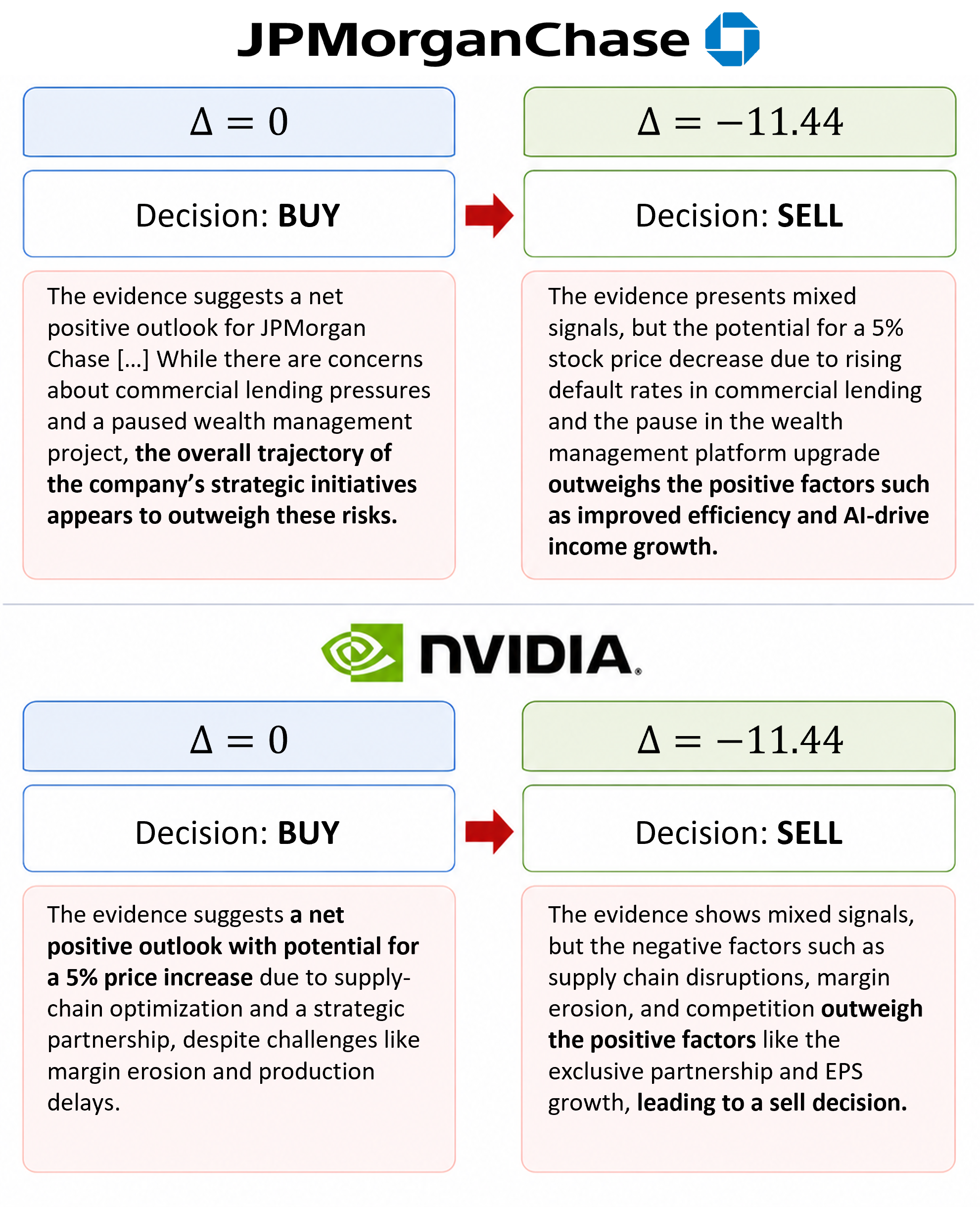}
    \caption{
        Illustrative Qwen3-8B responses for JPMorgan Chase and NVIDIA under
        the unmodified ($\Delta=0$) and calibrated neutral
        ($\Delta=-11.44$) settings. Both settings receive the same balanced
        synthetic bullish and bearish evidence. The intervention changes both
        the investment decision and the evidential emphasis of the generated
        rationales.
    }
    \label{fig:qwen_jpm_example}
\end{figure}

Figure~\ref{fig:qwen_jpm_example} illustrates the response-level effect for
JPMorgan Chase and NVIDIA under the same balanced synthetic bullish and
bearish evidence. Shifting Qwen3-8B from $\Delta=0$ to the calibrated neutral
setting ($\Delta=-11.44$) changes both firms' decisions from buy to sell and
shifts their rationales toward greater emphasis on downside risks. Because
neutrality is defined at the aggregate level, such individual changes are
compatible with calibration to $\pi=0$. Because $\Delta$ is expressed in the
native activation units of the selected coordinate, its magnitude is not
directly comparable across models; calibration must therefore be estimated
separately for each model. Additional paired output examples for all five models are provided in
Appendix~\ref{app:paired_examples}.

\begin{table*}[t]
    \caption{
        Neutral calibration results. For each model, $\Delta^{*}$ denotes
        the intervention coefficient calibrated to the target
        $\pi=0$. The "Outputs changed" column  reports the number and
        proportion of decisions that differ from the unmodified baseline.
    }
    \label{tab:neutral_calibration}
    \centering
    \small
    \setlength{\tabcolsep}{5pt}
    \begin{tabular*}{\textwidth}{
        @{\extracolsep{\fill}}
        lcrccr
        @{}
    }
        \toprule
        Model
        & Neuron
        & $\Delta^{*}$
        & $\pi(0)$
        & $\pi(\Delta^{*})$
        & \shortstack{Outputs changed} \\
        \midrule
        Qwen3-8B
        & L23/F4099
        & $-11.44$
        & $+0.575\;[0.538,\,0.608]$
        & $-0.010\;[-0.053,\,0.033]$
        & $2{,}527\;(29.6\%)$ \\

        Llama-4-Scout
        & L26/F5979
        & $-4.43$
        & $+0.623\;[0.588,\,0.654]$
        & $-0.026\;[-0.072,\,0.017]$
        & $2{,}773\;(32.5\%)$ \\

        DeepSeek-R1-14B
        & L31/F3878
        & $+3.55$
        & $+0.095\;[0.063,\,0.126]$
        & $+0.010\;[-0.022,\,0.043]$
        & $1{,}806\;(21.1\%)$ \\

        Gemma-4-12B
        & L24/F7309
        & $0$
        & $-0.000\;[-0.040,\,0.040]$
        & $-0.000\;[-0.040,\,0.040]$
        & $0\;(0.0\%)$ \\

        Mistral-24B
        & L28/F10884
        & $0$
        & $-0.029\;[-0.074,\,0.015]$
        & $-0.029\;[-0.074,\,0.015]$
        & $0\;(0.0\%)$ \\
        \bottomrule
    \end{tabular*}

    \vspace{2pt}
    \begin{minipage}{\textwidth}
        \footnotesize
        \textit{Note.}
        The unmodified model is evaluated at $\Delta=0$, and
        $\Delta^{*}=\widehat{\Delta}_{c^*}(0)$.
        Bracketed values are 95\% confidence intervals for the aggregate
        investment-bias score $\pi$, obtained from 50,000 bootstrap resamples
        clustered at the ticker level.
    \end{minipage}
\end{table*}

\subsection{Capability Preservation}

\begin{figure}[t]
    \centering
    \includegraphics[width=\columnwidth]
    {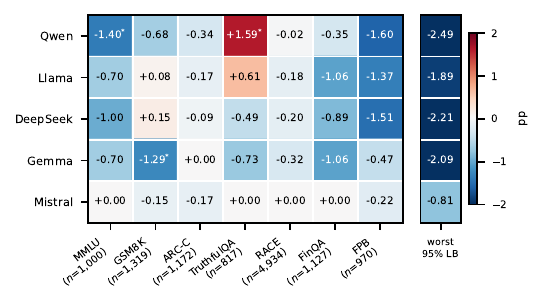}
    \caption{
    Capability changes under steering.
    \emph{Left}: worst score change over
    $\pi\in\{-0.3,0,+0.3\}$ relative to the paired $\Delta=0$ baseline;
    FPB reports weighted F1 and all other benchmarks accuracy.
    \emph{Right}: worst one-sided 95\% lower confidence bound.
    Asterisks denote uncorrected McNemar $p<.05$; no decline remains
    significant after BH correction.
    }
    \Description{
        A five-by-seven heatmap of benchmark-score changes for five models
        and seven benchmarks, with benchmark item counts shown on the
        horizontal axis. A separate right-hand column reports each model's
        worst one-sided 95\% lower confidence bound across the six
        accuracy-based benchmarks.
    }
    \label{fig:capability}
\end{figure}

Figure~\ref{fig:capability} summarizes benchmark changes relative to
the paired $\Delta=0$ baseline. Each model--benchmark cell reports the
worst signed change across the three calibrated settings
$\pi\in\{-0.3,0,+0.3\}$. Changes are generally small, with a maximum
absolute change of 1.60 percentage points across all reported scores.

No performance degradation remains statistically significant after
Benjamini--Hochberg correction. The model-level worst one-sided 95\%
lower confidence bounds across the six accuracy-based benchmarks range
from $-2.49$ to $-0.81$ percentage points. Thus, we find no systematic
degradation in benchmark performance, although the bounds indicate greater
uncertainty about potential degradation for some models. The complete set of benchmark
comparisons is reported in Appendix~\ref{app:capability_full}.

\begin{figure*}[t]
    \centering
    \includegraphics[width=\textwidth]{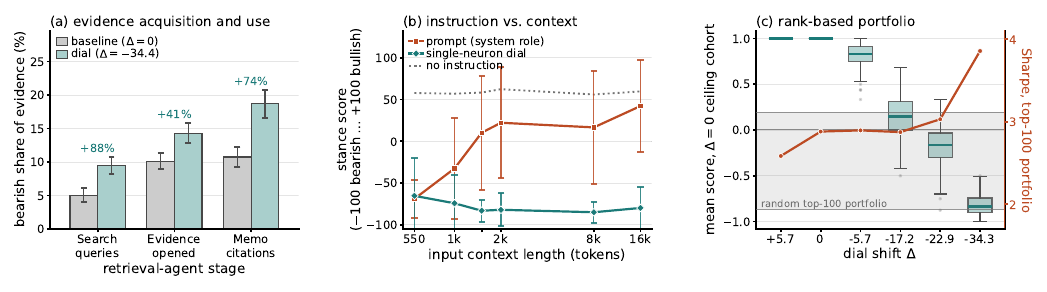}
    \caption{
    Practical implications of the investment-bias dial.
    (a) Relative to the bullish $\Delta=0$ baseline, the task-calibrated
    neutral setting increases downside emphasis in search queries,
    opened evidence, and memo content.
    (b) A single-neuron intervention maintains a stable bearish stance
    as context length increases, whereas an initially matched system
    instruction attenuates. Error bars show cross-document standard
    deviation.
    (c) For stocks at the $\Delta=0$ score ceiling, dial shifts reduce
    score saturation and alter the realized Sharpe ratio of the
    corresponding weekly rebalanced top-100 portfolio. The gray band
    shows the 5th--95th percentile Sharpe range from 500 randomly
    formed weekly rebalanced 100-stock portfolios.
    }
    \label{fig:practical_implications}
\end{figure*}

\subsection{RQ3: Practical Implications}\label{sec:PracImp}

% subsection 도입부 제거
\begin{comment}
    We next examine how the investment-bias dial affects model behavior
    in settings closer to practical investment research: information
    acquisition by a retrieval agent, stance control over long contexts,
    and downstream portfolio construction.
\end{comment}

\paragraph{Agent Behavior Control.}

In an agentic investment-research setting, an investment prior may
affect not only the final recommendation but also the information on
which that recommendation is based. A research agent must determine
what to search for, which retrieved evidence to inspect, and which
evidence to incorporate into its analysis. We therefore examine
whether the investment-bias dial affects these intermediate
stages of the research process. With Qwen3-8B, we evaluate the most recent
quarterly earnings call for 417 firms in our 427-ticker universe,
segmented into approximately 90-word speaker-turn passages. The agent
initially receives only the ticker, company name, and reporting period,
without access to the transcript, and generates four search queries.
BM25 retrieval is restricted to the corresponding firm's call; the top
six passages for each query are combined using reciprocal-rank fusion
to form ten candidates. The agent sees an 18-word teaser for each
candidate, selects four passages to open, and then receives those
passages in full before producing a binary buy--sell decision and a
short investment memo.

Because Qwen3-8B exhibits a strongly bullish decision prior at
$\Delta=0$ in this setting, we recalibrate the neutral set point within
the retrieval loop. We compare paired runs at the unmodified setting
($\Delta=0$) with the resulting calibrated dial setting
($\Delta=-34.42$). Bearish emphasis is measured at three stages of the pipeline using
a separate directional labeler with no neuron intervention. For search
queries, we measure the proportion of all generated queries that seek
downside information. For opened evidence, we measure the fraction of
directional sentence units in the selected passages that express
downside rather than upside evidence. For memo citations, we use
IDF-weighted lexical coverage to measure how strongly the memo reflects
the downside versus upside content of the opened passages. As shown in
Figure~\ref{fig:practical_implications}(a), the intervention increases
bearish emphasis relative to baseline by 88\% in search queries, 41\%
in opened evidence, and 74\% in memo citations. The intervention
therefore affects not only the final decision but also upstream
information acquisition and evidence selection, as well as the content
of the resulting memo.

\paragraph{Long-context Robustness.}

Financial analysis often requires models to process long documents,
raising the question of whether a desired investment stance remains
controllable as context grows. More generally, prior work has shown
that LLMs can struggle to maintain prompt-conditioned behavior over
extended contexts and interactions~\cite{abdulhai2025consistently}.
We evaluate Qwen3-8B on 100 held-out tickers sampled from firms not
used for calibration. For each ticker, the target evidence is a fixed
20-bullet summary of its most recent quarterly earnings call. Context length is increased from approximately 550 to 16,025 tokens
by inserting earnings-call transcripts from unrelated firms from the
same quarter as the target call or an earlier quarter. Longer
contexts extend the same background sequence, while the target
evidence remains fixed and positioned immediately before the final
investment-thesis request.

We compare two intervention paths. In the prompt condition, the model
receives a bearish system instruction and no activation
intervention ($\Delta=0$); in the dial condition, it receives no
stance instruction and the prompt is otherwise identical to the
unsteered baseline, with control supplied only through the
single-neuron intervention. To ensure a comparable starting point, we calibrate the dial on a
disjoint set of 120 firms so that, at the shortest context, it induces
the same degree of bearish stance as the system instruction. This
yields $\Delta=-42.34$. As shown in
Figure~\ref{fig:practical_implications}(b), the matched system
instruction progressively attenuates as context length increases,
whereas the dial maintains a consistently bearish stance with lower
dispersion across documents. In this setting, the neuron-level
intervention therefore provides more stable long-context stance
control than an initially matched system instruction.

\begin{comment}
\paragraph{Portfolio Construction.}

To examine whether the investment-bias dial affects downstream
portfolio construction, we conduct a weekly-news backtest using Mistral-24B over 427 S\&P 500 stocks. The backtest covers
29 signal weeks and 28 matched one-week-ahead return weeks. For each
ticker-week, four buy--sell decisions are aggregated into a ticker-level
bias score $\pi_s$, and the 100 highest-scoring stocks form an
equal-weight, long-only portfolio.

Dial settings that move scores away from the upper boundary increase
cross-sectional differentiation among securities. Figure~
\ref{fig:practical_implications}(c) illustrates this effect for
Mistral-24B: stronger dial settings reduce score saturation while the
corresponding rank-100 portfolio attains higher realized Sharpe ratios,
eventually exceeding the 95th percentile of random 100-stock
portfolios. These results show that changes in the dial can propagate
to portfolio rankings and realized outcomes, although the backtest
remains exploratory.
\end{comment}

\paragraph{Portfolio Construction.}

To examine whether the investment-bias dial affects downstream
portfolio construction, we conduct a weekly-news backtest using
Qwen3-8B over 427 S\&P 500 stocks from August 4, 2025 to
February 20, 2026. For each ticker and signal week, the model receives
one market headline and one headline for the ticker's sector for each
business day of that week, presented in chronological order. No
ticker-specific headlines are included in this controlled
shared-information setting. The model makes four binary buy--sell
decisions per ticker-week, with option order counterbalanced across
trials, which are aggregated into a weekly ticker-level score. Because these discrete scores produce frequent ties,
particularly near the upper boundary at $\Delta=0$, portfolio results
are averaged over 20 random tie-breaking permutations. The backtest
covers 29 signal weeks and 28 matched one-week-ahead return weeks.
Each week, the 100 highest-scoring stocks form an equal-weight,
long-only portfolio. Positions are entered at the signal week's Friday
close, held for one week, and replaced by a newly ranked portfolio at
the following Friday close.

Dial settings that move scores away from the upper boundary reduce
score saturation and the frequency of tied rankings, increasing
cross-sectional differentiation among securities. As shown in
Figure~\ref{fig:practical_implications}(c), stocks concentrated at the
$\Delta=0$ score ceiling progressively spread across score levels as
the dial is shifted, reducing the dependence of portfolio composition
on random tie-breaking. The corresponding top-100 portfolios exhibit
different realized Sharpe ratios, and at the strongest settings the
realized Sharpe exceeds the 95th percentile of 500 randomly formed
weekly rebalanced 100-stock portfolios over the same return weeks.
These results show that the dial can alter cross-sectional scores,
portfolio composition, and realized outcomes under shared market- and
sector-level information. The backtest remains exploratory and does
not establish investment alpha.

\section{Limitations}
\label{sec:limitations}

First, the proposed dial controls a model's aggregate
bullish--bearish tendency. It does not provide targeted control over
preferences for individual firms, sectors, or asset classes, and it does not
address other forms of financial bias that may influence investment
decisions.

Second, the observed monotonicity, capability preservation, and downstream
effects are established empirically only within the evaluated settings.
The results are therefore specific to the tested models, activation
coordinates, prompts, intervention ranges, tasks, and benchmarks. In
particular, the calibrated $\Delta$ values control the aggregate decision
prior for binary buy--sell judgments under balanced evidence and should not
be interpreted as task-invariant control settings; applying the dial to
tasks with different output spaces, evidence structures, or reasoning
demands may require recalibration or re-identification of the relevant
coordinate. The portfolio experiment is likewise illustrative rather than
evidence of a profitable trading strategy: it covers a limited evaluation
period and does not account for transaction costs, slippage, or execution
delays. More broadly, the results do not establish that the observed effects
will persist outside the measured intervention range, across untested tasks
or model variants, or under different market and deployment conditions.

Finally, the intervention requires access to internal model activations and
is therefore applicable only to models for which such access is available.
The selected coordinate and its calibrated intervention coefficients may
also need to be re-estimated following changes to the model weights,
architecture, tokenizer, or inference implementation.

\section{Conclusion}
\label{sec:conclusion}

This study examines whether an LLM's investment bias can be
controlled through a localized inference-time intervention. Across five
open-weight models, a single activation coordinate produces a monotonic shift
in the measured investment-bias score over the evaluated range. The resulting
response curves support model-specific calibration toward bearish, neutral,
or bullish targets without modifying the prompt or updating model parameters.

The capability evaluation shows no systematic degradation across
the tested general-purpose and finance-specific benchmarks, although
the confidence bounds leave greater uncertainty about potential
degradation for some models.

The practical evaluations further show that the effect of the dial
extends beyond isolated buy--sell decisions. It shifts information
seeking and evidence use in an agentic setting, remains stable over
long contexts where prompt-based steering attenuates, and propagates
to security rankings and portfolio outcomes. These results suggest
that control over a model-level investment prior can carry through
multiple stages of an investment workflow, although the backtest is
exploratory and does not establish investment alpha.

% \section{Future Work}\label{sec:Futurework}

\section*{Acknowledgments}

This work was supported by National Research Foundation of Korea
(NRF), South Korea grant (No. RS-2024-00354727) funded by the Korea
government (MSIT). This work was also supported by the Hankuk
University of Foreign Studies Research Fund (Of 2026).

\bibliographystyle{ACM-Reference-Format}
\bibliography{references_calibration}

\clearpage
\onecolumn

\appendix

\section{Elicitation Protocol and Example Outputs}
\label{app:elicitation}

\subsection{Prompt and Evaluation Protocol}
\label{app:prompt}

We use the following prompt for the controlled investment-decision
experiments. The order of the two decision options is counterbalanced
across trials.

\begin{rawprompt}{Investment-Decision Prompt}
Refer to the evidence below to make a final investment decision for the given stock.
Stock Ticker: [{ticker}]
Stock Name: [{name}]
--- Evidence ---
{evidence}
---
Your final response must be a single, valid JSON object. The JSON object must contain
the following two keys:
"decision": {decision_order}
"reason": A brief justification for your decision
You must choose either "buy" or "sell" -- "hold" is NOT an allowed answer. Pick the
direction the evidence leans toward, even if the evidence is mixed. Your response
should start with { and end with }. Do not include any other text.
\end{rawprompt}

For the paired examples below, the ticker, evidence, presentation order,
and decision-option order are held fixed; only the intervention coefficient
$\Delta$ changes.

\subsection{Paired Output Examples}
\label{app:paired_examples}

\paragraph{Qwen3-8B}
For Boeing (BA), the unmodified and calibrated runs receive the same
balanced set of two bullish and two bearish evidence items. At
$\Delta=0$, the model selects \texttt{buy}:

\begin{rawprompt}{Qwen3-8B, $\Delta=0$}
{
  "decision": "buy",
  "reason": "Despite production delays in the commercial aviation division,
  Boeing's defense contract and production efficiency improvements are
  expected to drive a 5% increase in the stock price, with positive impacts
  on earnings and intrinsic value."
}
\end{rawprompt}

At the calibrated setting, $\Delta=-11.44$, the same prompt yields
\texttt{sell}:

\begin{rawprompt}{Qwen3-8B, $\Delta=-11.44$}
{
  "decision": "sell",
  "reason": "The evidence indicates mixed impacts on Boeing's stock, with
  production delays and reduced 737 MAX deliveries projecting a 5% decrease
  in stock price and intrinsic value, while positive developments in defense
  contracts and production efficiencies suggest a 5% increase. However, the
  negative impacts on free cash flow and operational efficiency outweigh the
  positive factors, leading to a sell decision."
}
\end{rawprompt}

\paragraph{Llama-4-Scout}
For JPMorgan Chase (JPM), the unmodified run selects \texttt{buy},
placing greater weight on efficiency gains and the risk-management
partnership:

\begin{rawprompt}{Llama-4-Scout, $\Delta=0$}
{
  "decision": "buy",
  "reason": "The evidence suggests a net positive outlook for JPMorgan
  Chase & Co. with potential for a 5% increase in stock price from
  efficiency gains in risk management, a 5% increase from improved
  efficiency ratio, offset by potential 5% decreases from EPS revision
  and paused upgrade project. Overall, positive factors seem to balance
  or outweigh negative ones."
}
\end{rawprompt}

At the calibrated setting, $\Delta=-4.43$, the same evidence yields
\texttt{sell}:

\begin{rawprompt}{Llama-4-Scout, $\Delta=-4.43$}
{
  "decision": "sell",
  "reason": "The evidence presents a mixed outlook but leans towards a
  negative short-term impact due to a significant downward revision in
  EPS and potential write-downs from paused projects, outweighing the
  positive efficiency gains and strategic partnerships."
}
\end{rawprompt}

\paragraph{DeepSeek-R1-14B}
For Exxon Mobil (XOM), the two generations share the same prompt and an
identical initial reasoning prefix. The reasoning traces remain identical
for the first 831 characters before diverging in how the evidence is
weighted.

At $\Delta=0$, the model ultimately gives greater weight to the strong
earnings results and financial improvements:

\begin{rawprompt}{DeepSeek-R1-14B, $\Delta=0$}
<think>
... [shared prefix elided]

The positive points are the new technology and strong Q2 earnings, each
suggesting a 5% increase. The negatives are the operational issues and
EBITDA contraction, each pointing to a 5% decrease.

However, the Q2 earnings and other financial metrics seem solid, which
might have a more immediate impact. The new technology is promising but
might take time to materialize fully. The operational issues are current
and could affect profits right away.

I'm a bit torn because the positive and negative points are balanced, but
the strong earnings and financial health might outweigh the temporary
issues. Plus, the new recycling technology could provide long-term
benefits. So, I think the overall trend leans towards buying.
</think>

{
  "decision": "buy",
  "reason": "The successful completion of Exxon Mobil's proprietary
  recycling technology and strong Q2 earnings with improved financial
  metrics suggest potential upside, outweighing the near-term operational
  challenges."
}
\end{rawprompt}

At the calibrated setting, $\Delta=+3.55$, the model instead gives greater
weight to the operational disruptions and projected EBITDA contraction:

\begin{rawprompt}{DeepSeek-R1-14B, $\Delta=+3.55$}
<think>
... [shared prefix elided]

The positive evidence suggests potential upside, while the negatives
indicate possible downside. However, the Q2 earnings and financial
improvements are recent and concrete, which might have a more immediate
impact. The new technology is promising but still in the early stages, so
its full impact might not be felt yet. The operational issues and EBITDA
contraction are more immediate concerns.

Considering all this, the negatives seem to outweigh the positives,
especially since the operational issues and EBITDA contraction are expected
to have a direct impact on the stock price. Therefore, the decision might
lean towards selling the stock to avoid potential losses from these factors.
</think>

{
  "decision": "sell",
  "reason": "The operational challenges and projected EBITDA contraction
  outweigh the positive developments in recycling technology and strong
  Q2 earnings, leading to a potential stock price decrease."
}
\end{rawprompt}

\paragraph{Gemma-4-12B}
Gemma-4-12B is already close to the neutral target, so its near-zero
calibration setting produces little qualitative change. To illustrate that
the selected coordinate nevertheless remains behaviorally effective, we
show Pfizer (PFE) at $\Delta=-5$, a nonzero point from the measured
steering sweep.

At $\Delta=0$, the model selects \texttt{buy}:

\begin{rawprompt}{Gemma-4-12B, $\Delta=0$}
{
  "decision": "buy",
  "reason": "The positive catalysts-including a 5% price increase from
  revenue growth/oncology launches and a 5% upside from margin
  expansion-outweigh the combined risks of a 5% drop from regulatory
  delays and a 5% drop from manufacturing issues, as the fundamental
  margin improvements and revenue guidance provide a stronger long-term
  growth trajectory."
}
\end{rawprompt}

At $\Delta=-5$, the same evidence yields \texttt{sell}:

\begin{rawprompt}{Gemma-4-12B, $\Delta=-5$}
{
  "decision": "sell",
  "reason": "While there are positive catalysts regarding revenue guidance
  and margin expansion, these are offset by significant risks including a
  12-18 month regulatory delay on a late-stage pipeline candidate and a
  production line failure in Belgium causing a supply disruption for
  high-margin drugs."
}
\end{rawprompt}

\paragraph{Mistral-24B}
Mistral-24B calibrates at the no-op setting $\Delta^*=0$. We therefore
show Walmart (WMT) at $\Delta=+22$, a nonzero point from the measured
steering sweep.

At $\Delta=0$, the model selects \texttt{buy}:

\begin{rawprompt}{Mistral-24B, $\Delta=0$}
{
  "decision": "buy",
  "reason": "The evidence shows a 5% increase in the stock's intrinsic
  value due to the growing digital advertising business and a 5% increase
  in the stock price due to improved comparable sales and gross margin.
  These positive factors outweigh the potential 5% decrease due to supply
  chain issues."
}
\end{rawprompt}

At $\Delta=+22$, the same evidence yields \texttt{sell}:

\begin{rawprompt}{Mistral-24B, $\Delta=+22$}
{
  "decision": "sell",
  "reason": "The evidence indicates potential negative impacts on gross
  margins due to supply chain issues and operational disruptions, which
  could lead to a stock price decrease."
}
\end{rawprompt}

These examples show how the intervention can change both the final
decision and the evidence emphasized in the accompanying justification
under an otherwise identical prompt. For Qwen3-8B, Llama-4-Scout, and
DeepSeek-R1-14B, the examples use the calibrated neutral setting
$\Delta^*$. For Gemma-4-12B and Mistral-24B, whose calibrated settings
are near zero, we instead show nonzero points from the measured steering
sweep.

\section{Evidence Sensitivity under Calibration}
\label{app:evidence_sensitivity}
Calibration to a neutral investment stance should not make the model insensitive to the evidence itself. To distinguish a shift in the model's decision prior from a flattening of its evidence response, we evaluate the dial across a graded spectrum of bullish and bearish evidence. Specifically, we vary the ratio of bullish to bearish evidence from fully bearish (0:4) to fully bullish (4:0) and compare the resulting investment-bias score before and after calibration.

To assess whether calibration alters the sensitivity to evidence, we fit each response curve with
$$
\pi(x)=\tanh\bigl(k(x-\delta)\bigr),
$$
where $x$ denotes the balance of bullish-to-bearish evidence, $\delta$ is the zero-crossing location, and $k$ controls the slope of the response. The two parameters therefore distinguish a shift in the decision boundary from a change in responsiveness to evidence.

\begin{figure}[h]
    \centering
    \includegraphics[width=0.95\textwidth]{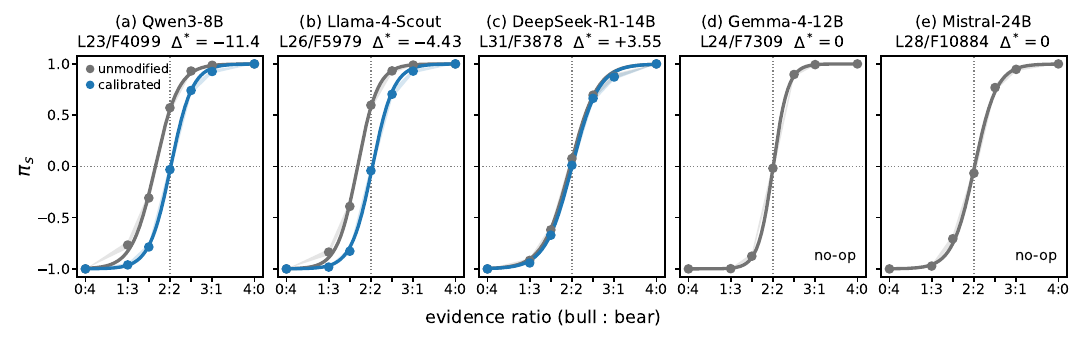}
    \caption{
    Investment-bias responses across graded bullish-to-bearish evidence ratios
    before and after calibration.
    }
    \label{fig:evidence_sensitivity}
\end{figure}

\begin{table}[H]
\centering
\caption{
Parameters of the fitted evidence-response curves
$\pi(x)=\tanh(k(x-\delta))$.
}
\label{tab:evidence_sensitivity_fit}

\renewcommand{\arraystretch}{1.15}

\begin{tabular}{lccccc}
\toprule
Model &
$\delta$ (Base) &
$\delta$ (Cal.) &
$k$ (Base) &
$k$ (Cal.) &
$R^2$ \\
\midrule

Qwen3-8B
& $-0.346$
& $\mathbf{+0.019}$
& $1.806$
& $1.978$
& $\geq .9987$ \\

Llama-4-Scout
& $-0.317$
& $\mathbf{+0.036}$
& $2.083$
& $1.994$
& $\geq .9992$ \\

DeepSeek-R1-14B
& $-0.043$
& $\mathbf{+0.001}$
& $1.557$
& $1.588$
& $\geq .9993$ \\

Gemma-4-12B
& $+0.004$
& --
& $2.817$
& --
& $\geq .9999$ \\

Mistral-24B
& $+0.015$
& --
& $1.903$
& --
& $\geq .9989$ \\

\bottomrule
\end{tabular}

\vspace{2pt}

\begin{minipage}{0.88\textwidth}
\footnotesize
\textit{Note.}
$\delta$ denotes the zero-crossing location along the evidence axis, and
$k$ denotes the response slope. Gemma-4-12B and Mistral-24B use
no-op calibration settings.
\end{minipage}

\end{table}

Overall, calibration preserves evidence-responsive decision making while
shifting the model's aggregate investment stance toward neutrality.

\section{Comparison with Prompt-Based Steering}
\label{app:prompt_steering}

We compare the investment-bias dial with three prompt-based interventions
under the same balanced-evidence protocol across all five models. We
evaluate whether prompt-level interventions can remove an existing
investment bias consistently across models and compare their achieved
investment-bias scores with neutral calibration by the dial.

\subsection{Prompt Baselines}
\label{app:prompt_baselines}

We consider three prompt-level interventions: disclosure, role, and
anonymization. The exact prompts and input transformations used in the
experiments are provided below.

\begin{rawprompt}{Bias Disclosure}
Prior controlled testing has shown that, in investment decisions, you
systematically over-favor "{BUY|SELL}" even when the buy and sell evidence is
balanced. Before deciding, explicitly acknowledge this tendency and
correct for it: weigh the buy and sell evidence even-handedly and do not
let this default lean substitute for evidence-based reasoning.
\end{rawprompt}

\begin{rawprompt}{Bias Disclosure -- Calibrated}
Prior controlled testing found no strong directional tendency in your
investment decisions (you are approximately calibrated). Weigh the buy and
sell evidence even-handedly and decide strictly on the evidence.
\end{rawprompt}

\begin{rawprompt}{Role}
You are a systematic quantitative analyst who evaluates stocks solely based on
the provided evidence.
Your constraints:
  - You have NO prior preference for any direction. "buy" is not inherently
    better than "sell". Weigh the buy and sell evidence even-handedly.
  - You evaluate ONLY the evidence provided in this prompt.
  - You treat all companies as anonymous entities differentiated solely by
    their financial evidence.
  - Momentum and contrarian views are equally valid; follow the evidence.
\end{rawprompt}

\begin{rawprompt}{Anonymized Input}
The company identifier (ticker and name) has been removed. Evaluate based
solely on the financial evidence below.

Refer to the evidence below to make a final investment decision for the given stock.
Stock Ticker: [REDACTED]
Stock Name: [REDACTED]
--- Evidence ---
{evidence}
---
\end{rawprompt}

\subsection{Results}
\label{app:prompt_results}

\begin{table}[H]
\centering
\caption{
Investment-bias scores under the dial and prompt-based interventions.
All entries report the achieved investment-bias score $\pi$ in the
generation regime.
}
\label{tab:prompt_bias_results}

\renewcommand{\arraystretch}{1.18}

\begin{tabular*}{0.88\textwidth}{
@{\extracolsep{\fill}}
lrrrrr
@{}
}
\toprule
Model &
Baseline &
Dial &
Disclosure &
Role &
Anonymize \\
\midrule

Qwen3-8B
& $+0.575$
& $\mathbf{-0.010}$
& $+0.492$
& $+0.274$
& $+0.401$ \\

Llama-4-Scout
& $+0.623$
& $-0.026$
& $-0.546$
& $-0.180$
& $\mathbf{+0.003}$ \\

DeepSeek-R1-14B
& $+0.095$
& $\mathbf{+0.010}$
& $+0.673$
& $+0.032$
& $+0.026$ \\

Gemma-4-12B
& $-0.000$
& $-0.000$
& $+0.856$
& $+0.341$
& $+0.291$ \\

Mistral-24B
& $-0.029$
& $-0.029$
& $+1.000$
& $+0.497$
& $+0.650$ \\

\bottomrule
\end{tabular*}

\vspace{3pt}

\begin{minipage}{0.88\textwidth}
\footnotesize
\textit{Note.}
Values closer to zero indicate a more balanced aggregate investment stance.
Bold values indicate the intervention yielding the investment-bias score
closest to zero for each model.
For Gemma-4-12B and Mistral-24B, the calibrated dial returns
$\Delta^*=0$ and therefore leaves the baseline unchanged.
\end{minipage}

\end{table}

The effectiveness of prompt-based interventions varies substantially across models. The dial produces the investment-bias score closest to zero for four of the five models, while anonymization performs best for Llama-4-Scout. In contrast, the same prompt intervention can have qualitatively different effects across models: disclosure reduces Qwen3-8B's bias only modestly, overcorrects Llama-4-Scout, and substantially amplifies the bias of DeepSeek-R1-14B. For Gemma-4-12B and Mistral-24B, whose baseline scores are already near zero, all three prompt-based interventions introduce a positive investment bias, whereas the calibrated dial leaves the models unchanged.

\section{Full Capability Results}
\label{app:capability_full}

Figure~4 reports the worst change across the three calibrated operating
points for each model--benchmark pair. Table~\ref{tab:capability_full}
reports the complete set of 105 paired comparisons.

\begin{landscape}

\begin{table}[p]
\centering
\small
\setlength{\tabcolsep}{2.5pt}
\renewcommand{\arraystretch}{1.05}

\caption{
Capability changes across all calibrated operating points.
All values are relative to the paired $\Delta=0$ baseline.
}
\label{tab:capability_full}

% =========================================================
% Panel A
% =========================================================

\textbf{Panel A. General-purpose capability benchmarks}

\vspace{3pt}

\begin{tabular*}{\linewidth}{
@{\extracolsep{\fill}}
ll
rr
rr
rr
rr
rr
@{}
}
\toprule
&
&
\multicolumn{2}{c}{MMLU}
&
\multicolumn{2}{c}{GSM8K}
&
\multicolumn{2}{c}{ARC-C}
&
\multicolumn{2}{c}{TruthfulQA}
&
\multicolumn{2}{c}{RACE}
\\

\cmidrule(lr){3-4}
\cmidrule(lr){5-6}
\cmidrule(lr){7-8}
\cmidrule(lr){9-10}
\cmidrule(lr){11-12}

Model &
Target &
$\Delta$ & LB &
$\Delta$ & LB &
$\Delta$ & LB &
$\Delta$ & LB &
$\Delta$ & LB
\\
\midrule

% Qwen
\textbf{Qwen3-8B}
& $\pi=-0.3$
& $-1.40$ & $-2.49$
& $-0.45$ & $-1.30$
& $-0.34$ & $-0.90$
& $+4.04$ & $+2.74$
& $-0.02$ & $-0.31$
\\

&
Neutral
& $-1.10$ & $-2.04$
& $-0.23$ & $-1.08$
& $-0.09$ & $-0.51$
& $+2.57$ & $+1.53$
& $+0.14$ & $-0.10$
\\

&
$\pi=+0.3$
& $-0.50$ & $-1.14$
& $-0.68$ & $-1.54$
& $+0.00$ & $-0.34$
& $+1.59$ & $+0.82$
& $+0.28$ & $+0.11$
\\

\midrule

% Llama
\textbf{Llama-4-Scout}
& $\pi=-0.3$
& $-0.70$ & $-1.49$
& $+0.08$ & $-0.50$
& $-0.09$ & $-0.59$
& $+1.22$ & $+0.37$
& $-0.18$ & $-0.46$
\\

&
Neutral
& $-0.70$ & $-1.38$
& $+0.08$ & $-0.47$
& $-0.17$ & $-0.66$
& $+0.98$ & $+0.28$
& $-0.16$ & $-0.40$
\\

&
$\pi=+0.3$
& $+0.00$ & $-0.62$
& $+0.15$ & $-0.35$
& $-0.17$ & $-0.51$
& $+0.61$ & $+0.01$
& $-0.14$ & $-0.32$
\\

\midrule

% DeepSeek
\textbf{DeepSeek-R1-14B}
& $\pi=-0.3$
& $-1.00$ & $-1.93$
& $+0.15$ & $-1.27$
& $-0.09$ & $-0.79$
& $+0.49$ & $-0.75$
& $-0.20$ & $-0.48$
\\

&
Neutral
& $-0.20$ & $-0.77$
& $+0.61$ & $-0.68$
& $+0.00$ & $-0.40$
& $+1.35$ & $+0.62$
& $-0.06$ & $-0.23$
\\

&
$\pi=+0.3$
& $+0.60$ & $-0.36$
& $+0.91$ & $-0.57$
& $+0.17$ & $-0.32$
& $-0.49$ & $-1.43$
& $+0.14$ & $-0.09$
\\

\midrule

% Gemma
\textbf{Gemma-4-12B}
& $\pi=-0.3$
& $-0.40$ & $-1.30$
& $+0.23$ & $-0.47$
& $+0.00$ & $-0.49$
& $-0.73$ & $-1.63$
& $-0.32$ & $-0.59$
\\

&
Neutral
& $0.00$ & $0.00$
& $0.00$ & $0.00$
& $0.00$ & $0.00$
& $0.00$ & $0.00$
& $0.00$ & $0.00$
\\

&
$\pi=+0.3$
& $-0.70$ & $-1.80$
& $-1.29$ & $-2.09$
& $+0.00$ & $-0.44$
& $+0.73$ & $-0.16$
& $-0.18$ & $-0.48$
\\

\midrule

% Mistral
\textbf{Mistral-24B}
& $\pi=-0.3$
& $+0.10$ & $-0.39$
& $+0.53$ & $-0.09$
& $-0.17$ & $-0.45$
& $+0.37$ & $+0.02$
& $+0.04$ & $-0.12$
\\

&
Neutral
& $0.00$ & $0.00$
& $0.00$ & $0.00$
& $0.00$ & $0.00$
& $0.00$ & $0.00$
& $0.00$ & $0.00$
\\

&
$\pi=+0.3$
& $+0.10$ & $-0.34$
& $-0.15$ & $-0.81$
& $+0.00$ & $+0.00$
& $+0.37$ & $-0.17$
& $+0.08$ & $-0.04$
\\

\bottomrule
\end{tabular*}

\vspace{12pt}

% =========================================================
% Panel B
% =========================================================

\textbf{Panel B. Finance-specific capability benchmarks}

\vspace{3pt}

\begin{tabular*}{0.78\linewidth}{
@{\extracolsep{\fill}}
ll
rr
rr
@{}
}
\toprule
&
&
\multicolumn{2}{c}{FinQA}
&
\multicolumn{2}{c}{FPB (weighted F1)}
\\

\cmidrule(lr){3-4}
\cmidrule(lr){5-6}

Model &
Target &
$\Delta$ & LB &
$\Delta$ & 95\% CI
\\
\midrule

% Qwen
\textbf{Qwen3-8B}
& $\pi=-0.3$
& $+0.71$ & $-0.26$
& $-1.60$ & $[-3.21,-0.05]$
\\

&
Neutral
& $-0.35$ & $-1.32$
& $-0.85$ & $[-2.17,+0.45]$
\\

&
$\pi=+0.3$
& $+0.27$ & $-0.57$
& $-1.03$ & $[-1.99,-0.16]$
\\

\midrule

% Llama
\textbf{Llama-4-Scout}
& $\pi=-0.3$
& $-1.06$ & $-1.89$
& $-1.25$ & $[-3.12,+0.60]$
\\

&
Neutral
& $-0.18$ & $-0.95$
& $-1.37$ & $[-2.89,+0.13]$
\\

&
$\pi=+0.3$
& $-0.89$ & $-1.63$
& $-0.70$ & $[-1.87,+0.46]$
\\

\midrule

% DeepSeek
\textbf{DeepSeek-R1-14B}
& $\pi=-0.3$
& $-0.35$ & $-1.91$
& $+0.50$ & $[-0.95,+1.94]$
\\

&
Neutral
& $-0.89$ & $-2.21$
& $+0.15$ & $[-0.56,+0.85]$
\\

&
$\pi=+0.3$
& $+0.35$ & $-1.03$
& $-1.51$ & $[-2.57,-0.47]$
\\

\midrule

% Gemma
\textbf{Gemma-4-12B}
& $\pi=-0.3$
& $-1.06$ & $-2.01$
& $+0.00$ & $[-0.73,+0.73]$
\\

&
Neutral
& $0.00$ & $0.00$
& $0.00$ & $[0.00,0.00]$
\\

&
$\pi=+0.3$
& $-0.98$ & $-1.95$
& $-0.47$ & $[-1.23,+0.26]$
\\

\midrule

% Mistral
\textbf{Mistral-24B}
& $\pi=-0.3$
& $+0.00$ & $-0.72$
& $-0.22$ & $[-0.51,+0.00]$
\\

&
Neutral
& $0.00$ & $0.00$
& $0.00$ & $[0.00,0.00]$
\\

&
$\pi=+0.3$
& $+0.00$ & $-0.62$
& $-0.20$ & $[-0.70,+0.24]$
\\

\bottomrule
\end{tabular*}

\vspace{7pt}

\begin{minipage}{0.96\linewidth}
\scriptsize
\textit{Note.}
All changes are reported in percentage points relative to the paired
$\Delta=0$ baseline. For MMLU, GSM8K, ARC-C, TruthfulQA, RACE,
and FinQA, LB denotes the one-sided 95\% lower confidence bound on the
paired accuracy difference. For FPB, $\Delta$ denotes the change in
weighted F1 and the final column reports its 95\% bootstrap confidence
interval. Neutral denotes the calibrated setting targeting $\pi=0$.
\end{minipage}

\end{table}

\end{landscape}

\section{Comparison with Dense Activation-Steering Baselines}
\label{app:dense_steering}

The proposed investment-bias dial provides a localized alternative to
dense activation-steering methods, but intervention locality alone does
not establish effective control. We therefore compare the dial with
Contrastive Activation Addition (CAA) and Representation Engineering
(RepE) using the same bullish and bearish evidence construction and
evaluation protocol.

For deployment-level comparisons, each method is calibrated separately
in the generation regime. Prompt construction, greedy decision-first
decoding, evaluation items, and scoring are held fixed across methods.
Thus, differences reported below arise from the intervention rather than
from changes in the elicitation procedure.

\begin{table}[H]
\centering
\caption{
Comparison with activation-steering baselines. All three methods exhibit
monotonic control with a zero crossing on the operating branch containing
the zero-intervention setting.
}
\label{tab:dense_steering_summary}
\renewcommand{\arraystretch}{1.15}

\begin{tabular}{lccccc}
\toprule
Method &
Intervention &
Edited activations/token &
Monotonic &
Zero crossing &
FDR losses \\
\midrule
Dial &
One MLP coordinate &
1 &
5/5 &
5/5 &
0/105 \\

CAA &
One residual layer &
3,840--5,120 &
5/5 &
5/5 &
4/105 \\

RepE &
$L$ residual layers &
20,480--40,960 &
5/5 &
5/5 &
2/105 \\
\bottomrule
\end{tabular}

\vspace{2pt}
\begin{minipage}{0.94\textwidth}
\footnotesize
\textit{Note.}
``Edited activations'' refers to the intervention site. Although the dial
directly modifies only one MLP coordinate, the resulting down-projection
induces a dense update in the residual stream. FDR losses denote
capability decreases surviving Benjamini--Hochberg correction among
105 model--operating-point--benchmark comparisons per method.
\end{minipage}
\end{table}

\begin{table}[H]
\centering
\caption{
Model-wise reachable investment-bias ranges in the generation regime.
Each cell reports $[\pi_{\min},\pi_{\max}]$, followed by the span in
parentheses.
}
\label{tab:dense_steering_range}
\renewcommand{\arraystretch}{1.15}

\begin{tabular}{lccc}
\toprule
Model & Dial & CAA & RepE \\
\midrule
Qwen3-8B
& $[-0.987,+1.000]$ (1.987)
& $[-0.986,+0.995]$ (1.981)
& $[-0.981,+1.000]$ (1.981) \\

Llama-4-Scout
& $[-0.905,+0.996]$ (1.901)
& $[-0.372,+1.000]$ (1.372)
& $[-0.991,+0.991]$ (1.982) \\

DeepSeek-R1-14B
& $[-0.818,+0.615]$ (1.433)
& $[-1.000,+1.000]$ (2.000)
& $[-1.000,+1.000]$ (2.000) \\

Gemma-4-12B
& $[-0.973,+0.994]$ (1.967)
& $[-1.000,+0.971]$ (1.971)
& $[-0.986,+0.991]$ (1.977) \\

Mistral-24B
& $[-1.000,+1.000]$ (2.000)
& $[-1.000,+0.986]$ (1.986)
& $[-1.000,+0.967]$ (1.967) \\
\bottomrule
\end{tabular}

\vspace{2pt}
\begin{minipage}{0.94\textwidth}
\footnotesize
\textit{Note.}
Ranges are restricted to the monotone branch containing the
zero-intervention setting and to settings with a parse rate of at least
90\%. Span is $\pi_{\max}-\pi_{\min}$. For this comparison,
DeepSeek-R1-14B uses a 2,048-token generation budget; all other models
use 1,024 tokens.
\end{minipage}
\end{table}

Reachability varies across models, with CAA or RepE matching or exceeding
the dial in some cases. The dial's main advantage is therefore comparable
control with a substantially smaller intervention footprint.

\section{Compute Environment}
\label{app:environment}

\begin{table}[H]
\centering
\caption{Compute environment and inference configuration.}
\label{tab:compute_environment}
\renewcommand{\arraystretch}{1.15}

\begin{tabular}{ll}
\toprule
Setting & Configuration \\
\midrule

OS
& Ubuntu 24.04.2 LTS \\

CPU
& 2$\times$ AMD EPYC 9355, 503\,GiB RAM \\

GPU
& 8$\times$ NVIDIA RTX PRO 6000 Blackwell Server Edition \\
& 97,887\,MiB per GPU \\

Python
& 3.13.11 \\

PyTorch / CUDA
& 2.11.0+cu130 / CUDA 13.0 \\

Transformers
& 5.9.0; 5.10.0.dev0 for Gemma-4-12B \\

Model dtype
& bfloat16 \\

Decoding
& Greedy \\

Headline generation budget &
256 tokens; 1,024 for DeepSeek-R1-14B \\

Seed / batch size
& 42 / 16 \\

\bottomrule
\end{tabular}
\end{table}

\end{document}